\documentclass[11pt]{article}
\usepackage{coling2020}
\usepackage{times}
\usepackage{url}
\usepackage{latexsym}

\usepackage{graphicx}
\usepackage{booktabs} 
\usepackage{multirow}

\usepackage{amsmath}
\usepackage{amssymb}
\usepackage{wrapfig}

\usepackage{verbatim}
\usepackage[normalem]{ulem}

\newcommand{\citet}{\newcite}
\newcommand{\citep}{\cite}

\colingfinalcopy 

\title{Keep it Consistent: Topic-Aware Storytelling from an Image Stream via Iterative Multi-agent Communication}

\author{Ruize Wang\textsuperscript{\rm 1}, Zhongyu Wei\textsuperscript{\rm 2,\rm3}\thanks{~~~Corresponding author}, 
Ying Cheng\textsuperscript{\rm 1}, Piji Li\textsuperscript{\rm 4}, \\ \textbf{ Haijun Shan\textsuperscript{\rm 5}, Ji Zhang\textsuperscript{\rm 5}, 
Qi Zhang\textsuperscript{\rm 6}, 
Xuanjing Huang\textsuperscript{\rm 6}}\\ 
\textsuperscript{\rm 1}	Academy for Engineering and Technology, Fudan University, China\\
\textsuperscript{\rm 2}School of Data Science, Fudan University, China\\ 
\textsuperscript{\rm 3}Research Institute of Intelligent and Complex Systems, Fudan University, China\\
\textsuperscript{\rm 4}Tencent AI Lab, China, \textsuperscript{\rm 5}Zhejiang Lab, China\\ 
\textsuperscript{\rm 6}School of Computer Science, Fudan University, China\\
\{rzwang18,zywei,chengy18,qz,xjhuang\}@fudan.edu.cn; \\ lipiji.pz@gmail.com; shanhaijun@zhejianglab.com; Ji.Zhang@usq.edu.au
}

\date{}

\begin{document}
\maketitle
\begin{abstract}
Visual storytelling aims to generate a narrative paragraph from a sequence of images automatically. Existing approaches construct text description independently for each image and roughly concatenate them as a story, which leads to the problem of generating semantically incoherent content. In this paper, we propose a new way for visual storytelling by introducing a topic description task to detect the global semantic context of an image stream. A story is then constructed with the guidance of the topic description. In order to combine the two generation tasks, we propose a multi-agent communication framework that regards the topic description generator and the story generator as two agents and learn them simultaneously via iterative updating mechanism. We validate our approach on VIST dataset, where quantitative results, ablations, and human evaluation demonstrate our method's good ability in generating stories with higher quality compared to state-of-the-art methods.

\end{abstract}

\section{Introduction}

Image-to-text generation is an important topic in artificial intelligence (AI) which connects computer vision (CV) and natural language processing (NLP). Popular tasks include image captioning~\citep{karpathy2015deep,ren2017deep,vinyals2017show} and question answering~\citep{Antol2015VQAVQ,Yu2017vqa,singh2019vqa}, aiming at generating a short sentence or a phrase conditioned on certain visual information. With the development of deep learning and reinforcement learning models, recent years witness promising improvement of these tasks for single-image-to-single-sentence generation. 

Visual storytelling moves one step further, extending the input and output dimension to a sequence of images and a sequence of sentences. It requires the model to understand the main idea of an image stream and generate coherent sentences. Most of existing methods~\citep{huang2016visual,liu2016storytelling,yu2017hierarchically,wang2018show,wang2020storytelling} for visual storytelling extend approaches of image captioning without considering topic information of the image sequence, which causes the problem of generating semantically incoherent content.

\begin{figure} [t]
\begin{center}
\includegraphics[width=0.6\textwidth]{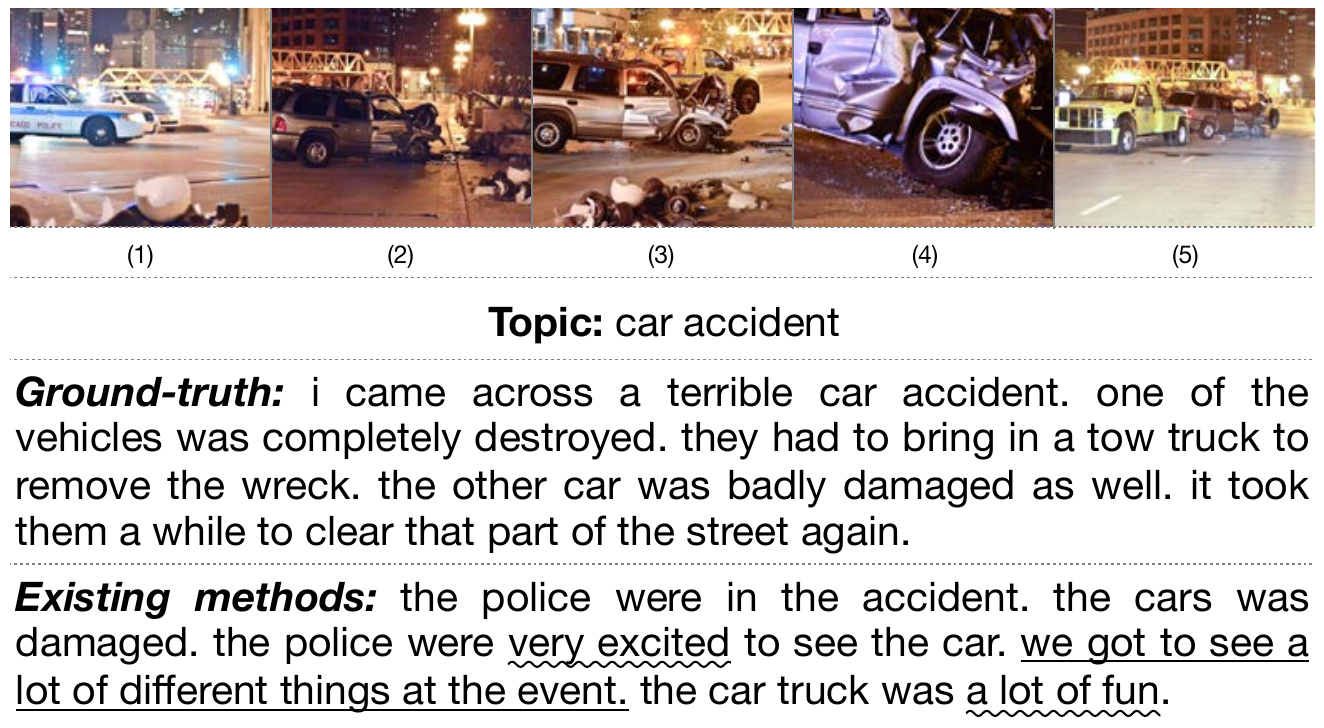}  
\end{center}
\vspace*{-3ex}
   \caption{An example of visual storytelling from VIST dataset consisting of five images with the topic of car accident. Two stories presented are from an automatic approach and a human annotator respectively. \uwave{Wavy lines} highlight the inappropriate sentiment of the machine-generated sentence. \underline{Underlines} indicate that the sentence provides little on-topic information about the image sequence.}
\label{fig:intro}
\vspace*{-2ex}
\end{figure}

An example of visual storytelling can be seen in Figure~\ref{fig:intro}. An image stream with five images about a car accident is presented accompanied with two stories. One is constructed by a human annotator and the other is produced by an automatic storytelling approach. There are two problems with the machine generated story. First, the sentiment expressed in the text is inappropriate. In face of a terrible car accident, the model uses some words with positive emotion, i.e., \textit{``excited''} and \textit{``fun''}. Second, some sentence is uninformative. The sentence \textit{``we got to see a lot of different things at the event''} provides little information about the car accident. This example shows that topic information about the image sequence is important for the story generator to produce an informative and semantically coherent story. 

In this paper, we introduce a novel task of topic description generation to detect the global semantic context of an image sequence and generate a story with the guidance of such topic information. In practice, we propose a framework named \textbf{T}opic-\textbf{A}ware \textbf{V}isual \textbf{S}tory \textbf{T}elling (\textbf{TAVST}) to tackle the two generation tasks: (1) topic description generation for the given image stream; (2) story generation with the guidance of the topic information. To effectively combine these two tasks, we propose a multi-agent communication framework that regards the topic description generator and the story generator as two agents. In order to enable the interaction of these two agents for information sharing, an iterative updating (IU) module is incorporated into the framework. Extensive experiments on the VIST dataset \citep{huang2016visual} show that our framework achieves better performance compared to state-of-the-art methods. Human evaluation also demonstrates that the stories generated by our model are better in terms of relevance, expressiveness and topic consistency.

\section{Approach}
\begin{figure*} [!t]
\centering
  \includegraphics[width=1\textwidth]{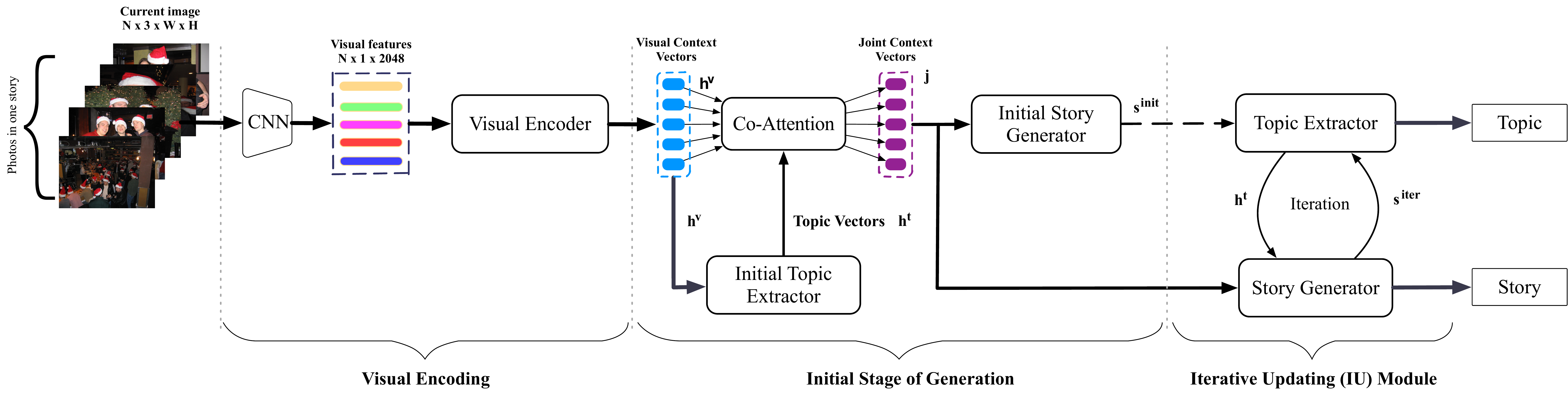}
  \caption{Overall architecture of our TAVST model. 
}
  \label{fig:over-view}
\end{figure*}

Given an image stream $x=(x_1,...,x_N)$, where $N$ is the number of images, we aim to output a topic description $y_{topic}$ and $N$ sub-stories to form a complete story $y=(y_1,...,y_N)$. The proposed framework mainly includes three stages, namely visual encoding, initial stage of generation and iterative updating (IU). The visual encoder (\S \ref{visual_encoder}) is employed to extract image features as visual context vectors. In the initial stage, we have the initial version of the two generation agents. The initial topic description generator (\S \ref{initial_topic_generator}) takes visual context vectors as input and generates a topic vector. The initial story generator (\S \ref{initial_story_generator}) combines the topic vector and visual context vectors via co-attention mechanism and construct the initial version of story. Considering that the generated story can also benefit the topic description generator, the two agents communicate with each other in the IU module (\S \ref{iu_module}) via message passing mechanism as fine tuning. The overall architecture of our proposed model is shown in Figure \ref{fig:over-view}. 
Each of these modules will be described in details in the following sections.

\subsection{Visual Encoder}
\label{visual_encoder}
Given an image stream $x$ with $N$ images, we first extract the high-level visual features $f_i$ of each image $x_i$ $(i\in1,...,N)$ through a CNN model. Then for the whole image stream, following the previous work \citep{Wang:2018tda}, a bidirectional gated recurrent unit (biGRU) is employed as the visual encoder:
\begin{equation}
	\setlength{\abovedisplayskip}{3pt} 
	\setlength{\belowdisplayskip}{3pt} 
\begin{split}\nonumber
f_i &= \text{ResNet}(x_i) \\ 
\overrightarrow{h_{i,t}} &= \overrightarrow{\mbox{GRU}}(f_i, \overrightarrow{h_{i,t-1}}) \\
\overleftarrow{h_{i,t}} &= \overleftarrow{\mbox{GRU}}(f_i, \overleftarrow{h_{i,t+1}}) \\
\end{split}
\end{equation}
where $\overrightarrow{h_{i,t}}$ is the forward hidden state at time step $t$ of $i$-th visual feature $f_i$, while the $\overleftarrow{h_{i,t}}$ is the backward one. At each time step, the visual features are sequentially fed into the visual encoder to obtain visual context vectors, which have integrated the visual information from the observed images. 
At last, fused with the original visual representation, the final visual context vector $h^v_i$ 
can be calculated as:
{
\setlength\abovedisplayskip{1pt}
\setlength\belowdisplayskip{1pt}
\begin{equation}
h^v_i =\text{ReLU}([\overleftarrow{h_i};\overrightarrow{h_i}]+W_ff_i)
\end{equation}
}
where $W_f$ is a projection matrix.
\subsection{Initial Topic Description Generator}
\label{initial_topic_generator}

Given the visual context vector extracted from the image sequence, we first learn to generate the topic description. In practice, all visual context vectors $h^v_i$ are concatenated and then fed into the initial topic description generator that employes a gated recurrent unit (GRU) decoder. 

The output $p^{init}_{topic}$ of this decoder is a sequence of probability distribution over the whole topic vocabulary $\mathbb{V}_t$. The training loss of initial topic description generator is the cross-entropy $L^{init}_{topic(mle)}$ between the generated description $p^{init}_{topic}$ and the ground-truth topic description $p_{topic}$.

Note that at each time step, the decoder produces the a hidden state $h^t_i$. Once the last topic hidden state $h^t_M$ is obtained, we concatenate all topic hidden states $h^t=[h^t_1,...,h^t_M]$, $M > 1$ as the topic memory, which are fed into the story generation module. 


\subsection{Initial Story Generator with Co-attention Network}
\label{initial_story_generator}

The initial story generator is responsible for generating the story with the guidance of the topic description constructed by the initial topic description generator.

\paragraph{Co-attention Encoding}
In order to combine both visual information and topic information for story generation, we adopt a co-attention mechanism~\citep{Jing2018OnTA} for context information encoding. The structure of the co-attention encoding module can be seen in the Appendix. 
Specifically, given visual context vectors $h^v$ and topic vectors $h^t$, the affinity matrix $C$ is calculated by
{
\begin{equation}
C = \text{tanh}({h^t}^TW_{b}h^v) 
\end{equation}}where $W_b$ is the weight parameter. After calculating this matrix, we compute attentions weights over the visual context vectors and the topic vectors via the following operations:
{
\begin{equation}
\begin{split}
H^v &= \text{tanh}(W_vh^v+(W_th^t)C) \\
H^t &= \text{tanh}(W_th^t+(W_vh^v)C^T) \\
a^v &= \text{softmax}(w^T_{hv}H^v) \\
a^t &= \text{softmax}(w^T_{ht}H^t)
\end{split}
\end{equation}}where $W_v$, $W_t$, $w^T_{hv}$, $w^T_{ht}$ are the weight parameters. Based on the attention weights, the visual and semantic attentions are calculated as the weighted sum of the visual context vectors and the topic vectors:
{
\begin{equation}
a^v_{att} = \sum_{n=1}^N a^v_n h^v_n, a^t_{att} = \sum_{m=1}^M a^t_m h^t_m
\end{equation}
}
At last, we concatenate the visual and semantic attentions as $[a^v_{att};a^t_{att}]$, and then use a fully connected layer $W_{fc}$ to obtain the joint context vector:
{
\begin{equation}
 j = W_{fc}[a^v_{att};a^t_{att}]
\end{equation}

}

\paragraph{Story Decoding} 
In the story decoding stage, each joint context vector $j_i$ is fed into a GRU decoder to generate a sub-story sentence $y_i$ for the corresponding image. 
Formally, the generation process can be written as:
{
\begin{align}
s^i_t &= \text{GRU}(s^i_{t-1}, [w^i_{t-1}, j_i])  \\
p(w^i_{t}|w^i_{1:t-1}) &= \text{softmax}(\text{MLP}(s^i_t))
\end{align}
}where $s^i_t$ denotes the $t$-th hidden state of $i$-th GRU. We concatenate the previous word token $w^i_{t-1}$ and the context vector $j_i$ as the input at each step. The output $p$ is a probability distribution over the whole story vocabulary $\mathbb{V}_s$. 

\paragraph{Loss Function for Training} 
We define two different loss functions including cross-entropy (MLE) and reinforce (RL). MLE loss is show in Equation~\ref{story_mle}:
{
\begin{equation}
\label{story_mle}
L^{init}_{story(mle)}(\theta_1) = -\sum^{T}_{t=1}log(p_{\theta_1}(y^*_t|y^*_1,...,y^*_{t-1}))
\end{equation}}where $\theta_1$ is the parameter of story generator; $y^*$ is the ground-truth story and $y^*_t$ denotes the $t$-th word in $y^*$. 

Recently, reinforcement learning has shown effectiveness for training text generation model via introducing automatic metrics (e.g., METEOR) to guide the training process~\citep{Wang:2018tda}. We also explore the RL-based approach to train our generator. The reinforcement learning (RL) loss can be written as:
{
\begin{equation}
\label{story_rl}
L^{init}_{story(rl)}(\theta_1) = -E_{y \sim p_{\theta_1}}(r(y; y^*)-b)^2
\end{equation}
}where $r$ is a sentence-level metric for the sampled sentence $y$ and the ground-truth $y^*$; $b$ is the baseline which can be an arbitrary function but a linear layer in our experiments for simply. To stabilize the RL training process, a simple way is to linearly combine MLE and RL objectives as follows \citep{wu2018study}:
{
\begin{equation}
\label{story_com}
L^{init}_{story(com)} = \alpha L^{init}_{story(rl)}+(1-\alpha)L^{init}_{story(mle)}
\end{equation}
}where hyper-parameter $\alpha$ is employed to control the trade-off between MLE and RL objectives.

In the initial stage, a combined loss function of $L^{init}_{story(com)}$ and $L^{init}_{topic}$ is computed through:
{
\begin{equation}
\label{init}
L^{init} = \lambda_1 L^{init}_{story(com)}+(1-\lambda_1)L^{init}_{topic(mle)}
\end{equation}
}where hyper-parameter $\lambda_1$ is employed to balance these losses.

\subsection{Iterative Updating Module}
\label{iu_module}

Considering that the generated story would also be helpful for the generation of topic description, we design an iterative updating module for the two agents to interact with each other and update iteratively. In IU module, we generate the topic description from the previously generated story, and then use such topic information to further guide story generation. To distinguish the two agents from those of initial version, we call them the IU version.

\paragraph{IU Topic Description Generator}
We envisage that the generated story is able to provide more accurate information for topic description generation than visual information. Therefore, instead of using visual information as input, the IU version of topic description generator takes the generated story as input. Specifically, the last hidden states $s_{iter}$ of the IU story generator is used as input. Note that the IU topic description generator is initialized as its initial version and keeps training with the same objective. 

\paragraph{IU Story Generator}
The IU story generator shares the structure and parameters with the initial story generator. It takes both topic vector and visual context vector as input. In the decoding process, the story $y$ is the concatenation of the sub-stories $y_i$ generated by IU story generator, and the last hidden states $s_{iter}=[s^1_{last},...,s^N_{last}]$ of the IU story generator will be passed to IU topic description generator for iterative updating.

\paragraph{Loss Function for Training}

The training losses of the IU story generator are similar to Eq.(\ref{story_mle},\ref{story_rl},\ref{story_com}), including the combination loss $L^{iter}_{story(com)}$ of $L^{iter}_{story(mle)}$ and $L^{iter}_{story(rl)}$. 

At each iteration stage $n$, the IU module loss $L^{iter}_n$ is the weighted sum of IU topic description generation loss $L^{iter}_{topic(mle)}$ and IU story generation loss $L^{iter}_{story(com)}$:
{
\setlength\abovedisplayskip{3pt}
\setlength\belowdisplayskip{3pt}
\begin{equation}
L^{iter}_n = \lambda_2 L^{iter}_{story(com)}+(1-\lambda_2)L^{iter}_{topic(mle)}
\end{equation}
}where hyper-parameter $\lambda_2$ is employed to balance these losses.

\subsection{Multi-Agent Training}
The IU topic description generator and IU story generator communicate with each other iteratively in the IU module until it reaches the given iteration number $N_{iter}$. 
The loss for IU module is:
{
\setlength\abovedisplayskip{1pt}
\setlength\belowdisplayskip{1pt}
\begin{equation}
L^{iter} = \sum_{n=1}^{N_{iter}}L^{iter}_n
\end{equation}
}

Therefore, to train the whole multi-agent learning framework, we introduce a combined loss $L$ which consists of the initial loss $L^{init}$ and IU module loss $L^{iter}$:
{
\setlength\abovedisplayskip{1pt}
\setlength\belowdisplayskip{1pt}
\begin{equation}
L = \beta L^{init}+(1-\beta)L^{iter}
\end{equation}
}where $\beta$ is a hyper-parameter to balance these losses. During training, our goal is minimizing $L$ using stochastic gradient descent. 


\section{Experiments}

\subsection{Datasets}
This paper utilizes VIST dataset~\citep{huang2016visual} for experiments. 
We use the same split settings as previous work \citep{Wang:2018tda}, inclduing 40,098/ 4,988/ 5,050 samples for training, validation and testing, respectively. Each sample (album) contains five images and a story consisting of five sentences. We use the title of each album as the ground-truth topic description.

\subsection{Implementation Details}
We use the pre-trained ResNet-152 \citep{he2016deep} to extract image features.
The vocabulary of story and topic includes words appearing no less than three times in the training set (i.e., story and title). We adopt GRU models for both visual encoder and other decoders, and the hidden size of them is 512. Expect the encoder is bidirectional, the other decoders are unidirectional. 
The batch size is set as 64 during the training. We use Adam \citep{kingma2015adam} with the initial learning rate of 0.0002.

We first pre-train the initial topic description generator using MLE. Then we pre-train both the topic description generator and the story generator jointly using MLE. The number of iteration $N_{iter}$ is set to 2, the weight of RL is $\alpha=0$, and hyper-parameters in loss optimization are set as $\lambda_1=0.7$, $\lambda_2=0.7$ and $\beta=0.3$, which are selected based on validation set \textbf{(the details about hyper-parameter sensitivity analysis are shown in Supplementary Material)}. After warm-up pre-training, $\alpha$ and learning-rate are set to 0.8 and 0.00002 to fine-tune using RL. Here we use METEOR scores as the reward. We select the best model which achieves the highest METEOR score on the validation set. The reason is that METEOR is proved to correlate better with human judgment than CIDEr-D in the small references case and superior to BLEU@N all the time \citep{vedantam2015cider,wang2018show}. During the test stage, we generate the stories by performing a beam-search with a beam size of 3.


\subsection{Models for Comparison}
We compare our proposed methods with several baselines for visual storytelling  as follows:

\textbf{seq2seq \citep{huang2016visual}:} It generates caption for each single model via classic sequence-to-sequence model and concatenate all captions to form the final story.
	
\textbf{h-attn-rank \citep{yu2017hierarchically}:} On top of the classic sequence-to-sequence model, it adds an additional RNN to select photos for story generation. 

\textbf{HPSR \citep{wang2019hierarchical}:} It introduces an additional RNN stacked on the RNN-based photo encoder to detect the scene change. Information from both RNNs are fed into an RNN for story generation. 

\textbf{AREL \citep{Wang:2018tda}:} It is based on the framework of reinforcement learning and the generation of a single word is treated as the policy. The reward model learns the reward function from human demonstrations.

\textbf{HSRL \citep{huang2019hierarchically}:} It is based on the framework of hierarchical reinforcement learning. The higher level agent is responsible for generating a local concept for each image as the guidance to the lower level agent for sentence generation . 

\textbf{VST:} This is the baseline version of our model without using topic information as guidance.

\textbf{TAVST w/o IU:} This is our proposed \emph{TAVST} method without IU module, which only equipped with initial topic description generator. 

\textbf{TAVST:} This is our full model. \emph{TAVST (MLE)} is trained using MLE loss, while \emph{TAVST (RL)} is trained via RL loss. 

{
	\begin{table*}[t!]
		\centering
		\begin{tabular}{lccccccc}
			\hline
			Methods & B-1&B-2&B-3&B-4&R-L&C&M\\
			\hline
			\multicolumn{7}{l}{\textbf{MLE}} \\
			\hline
			seq2seq \citep{huang2016visual} & $-$ & $-$ & $-$ & $3.5$ & $-$ & $6.8$ & 31.4\\
			h-attn-rank \citep{yu2017hierarchically} & $-$ & $-$ & 21.0 & $-$ & 29.5 & 7.5 & 34.1\\
			HPSR \citep{wang2019hierarchical} & 61.9 & 37.8 & 21.5  & 12.2 & \textbf{31.2} & 8.0 & 34.4\\
			\hline
			VST (MLE) & 62.3 & 38.0 & 21.8  & 12.7 & 29.7 & 7.8 & 34.3\\
            TAVST w/o IU (MLE)  & 63.1 & 38.6 & 22.9 & 14.0 & 29.7 & 8.5 & 35.1\\
            TAVST (MLE) & \textbf{63.6} & \textbf{39.3} & \textbf{23.4} & \textbf{14.2} & 30.3 & \textbf{8.7} & \textbf{35.4}\\
            \hline
			\multicolumn{7}{l}{\textbf{RL}} \\
			\hline
			AREL \citep{Wang:2018tda} & 63.7 & 39.0 & 23.1  & 14.0 & 29.6 & 9.5 & 35.0\\
			HSRL \citep{huang2019hierarchically} & - & - & -  & 12.3 & 30.8 & \textbf{10.7} & 35.2\\
			\hline
            TAVST w/o IU (RL)  & 63.5 & 39.2 & 23.2 & 14.3 & 30.0 & 8.7 & 35.3\\
			TAVST (RL) & \textbf{64.2} & \textbf{39.6} & \textbf{23.7} & \textbf{14.6} & \textbf{31.0} & 9.2 & \textbf{35.7}\\
			\hline
		\end{tabular}
		\caption{ Overall performance of story generation on VIST dataset for different models in terms of BLEU (B), METEOR (M), ROUGE-L (R-L), and CIDEr-D (C). \textbf{Bolded} numbers are the best performance in each column.}
		\label{table:automatic-eval}
		\vspace{-2mm}
	\end{table*}
}

\subsection{Automatic Evaluation Results}

We evaluate our model on two generation tasks i.e., story generation and topic description generation, in terms of four automatic metrics: BLEU \cite{papineni2002bleu}, ROUGE-L \citep{lin2004automatic}, METEOR \citep{banerjee2005meteor}, and CIDEr \citep{vedantam2015cider}.

\paragraph{Story Generation}
The overall experimental results are shown in Table \ref{table:automatic-eval}. \emph{TAVST (MLE)} outperforms all of the baseline models trained with MLE. This confirms the effectivness of topic information for generating better stories. Noticeably, compared with the RL-based models, our \emph{TAVST (MLE)} has already achieved a competitive performance and outperforms other RL models (i.e., \emph{AREL} and \emph{HSRL}) in terms of METEOR and BLEU@[2-4] metrics. After equipped with RL, our \emph{TAVST (RL)} model is able to further improve the performance, outperforming the two RL models in terms of all metrics except CIDEr-D. Our full model \emph{TAVST} (both MLE and RL versions) outperforms \emph{TAVST w/o IU}, which directly demonstrates the effectiveness of the IU module. \emph{TAVST w/o IU} achieves better performance than \emph{VST}, which proves that topic description generator can provide guidance for story generation.

\begin{wraptable}{r}{0.52\textwidth}
	\centering
		\vspace{-3.5mm}

	\begin{tabular}{lccccc}
		\hline
		Methods &B-1&B-2&R&C&M\\
		\hline
        TAVST w/o IU   & 5.1 & 2.2 & 12.6 &10.9 &4.8 \\
		TAVST  & \textbf{6.0} &\textbf{2.7} &\textbf{13.4} & \textbf{12.1} & \textbf{5.6}\\
		\hline
	\end{tabular}
	\smallskip
	\vspace{-3mm}

	\caption{ Performance of topic description generation.}
	\label{table:topic}
	\vspace{-2mm}
\end{wraptable}

\paragraph{Topic Description Generation}
Table \ref{table:topic} shows the results of topic description generation. \emph{TAVST} achieves higher performance compared to \emph{TAVST w/o IU}, indicating that the generated story is able to provide assistance for better topic description generation. In general, the description generator obtains low scores in terms of automatic metrics. observations on the dataset reveal that the length of titles for each album is relatively short, ranging from 2 to 6 words mostly. Given such a short reference, it is difficult for models to obtain high scores in terms of automatic metrics. We further look into the generated descriptions and some of them are actually semantically correct. For example, the reference is ``\textit{happy birthday party at my home}'' and the generated topic description is``\textit{the birthday gathering}''. Another example is that the reference is ``\textit{family feast}'' and the generated topic is ``\textit{dinner party}''. We believe such kind of topic description with similar meaning can still provide positive guidance for the story generator.

\subsection{Human Evaluation}
We perform two kinds of human evaluation through Amazon Mechanical Turk (AMT), namely the Turing test and the pairwise comparison. Since we only find one previous work~\citep{Wang:2018tda} which published the sampled results of their model, we chose it for comparison. In specific, we re-collect human labels for their sample results and stories generated by our models on the same sub-set of albums. A total of 150 stories (750 images) are used, and each of them is evaluated by 3 human evaluators. 

\begin{wraptable}{r}{0.48\textwidth}

\begin{center}
  \begin{tabular}{ l | c  c  c}
  \hline
    Method     & Win & Lose & Unsure \\
    \hline
    VST     & 22.4\% & 71.7\% & 5.9\% \\
    AREL    & 38.4\% & 54.2\% & 7.4\% \\
    \hline
    TAVST     & \textbf{42.7}\% & \textbf{52.2}\% & \textbf{5.1}\% \\ 
    \hline
  \end{tabular}
    \end{center}
    
    \label{table:turing}
    \smallskip
    \caption{Human evaluation results on Turing test.}

\end{wraptable}

\paragraph{Turing Test}
For the Turing test, we design a survey (as shown in Appendices) that contains an image stream, a generated story by our \emph{TAVST} model and a story written by a human. Evaluators are required to choose the story that is more likely written by a human. The experimental result (Table \ref{table:turing}) shows that 47.8\% of evaluators think the stories generated by our model are written by a human (v.s 38.4\% win rate from \emph{AREL}). 

\begin{table*}[!htp]

	\centering
	\small
	\begin{tabular}{c|ccc|ccc|ccc}
		\hline
		& \multicolumn{3}{c|}{TAVST  $vs$ VST} & \multicolumn{3}{c|}{TAVST  $vs$ AREL}& \multicolumn{3}{c}{TAVST  $vs$ GT} \\
		\hline
		Choice (\%) & TAVST  & VST & Tie & TAVST  & AREL & Tie & TAVST  & GT & Tie \\
		\hline
		Relevance & \textbf{56.00} & 42.00 & 2.00 & \textbf{51.56} & 47.56 & 0.89 & 41.44 & \textbf{48.20} & 10.36 \\  
		Expression & \textbf{60.00} & 38.44 & 1.56 & 46.22 & \textbf{53.33} & 0.44 & 42.12 & \textbf{56.31} & 1.58 \\ 
		Topic Consistency & \textbf{66.44} & 31.11 & 2.44 & \textbf{58.67} & 38.67 & 2.67 & 43.24 & \textbf{46.85} & 9.91 \\ 
		\hline
	\end{tabular}
	\vspace{-2mm}
\smallskip
	\caption{ Human evaluation results on pairwise comparison. }
	\label{table:human_pair}
	\vspace{-4mm}
\end{table*}

\paragraph{Pairwise Comparison}
A good story for an image stream should have three significant factors: (1) Relevance: the story should be relevant to the image stream. (2) Expressiveness: the story should be concrete and coherent, and have a human-like language style. (3) Topic Consistency: the story should be consistent to the topic. We compare our method with three other methods (\emph{VST}, \emph{AREL}, and \emph{ground-truth (GT)}) in terms of these three metrics.
In this annotation task, AMT evaluators need to compare two given stories according to these three factors and choose which story is better in terms of a certain factor. Results are shown in Table \ref{table:human_pair}. Our model performs better than the other two models in terms of relevance and topic consistency. The advantage of topic consistency is more promising. This proves that the topic description generator can help the story generation agent construct a more consistent story. 

\subsection{Further Analysis on Topic Consistency}
We further evaluate the quality of the generated story in terms of topic consistency from the perspective of sentiment. Specifically, we employ a lexicon-based approach using a subjectivity lexicon~\citep{wilson-etal-2005-recognizing}. We count the number of sentiment words in each sentence for the polarity evaluation. The score will be 1,0,-1 if a sentence is positive, neutral and negative, respectively. Based on the score for each sentence, two qualitative experiments are designed to measure the in-story sentiment consistency and topic-story sentiment consistency. 

\begin{figure}[t!]
\centering
  \includegraphics[width=0.6\textwidth]{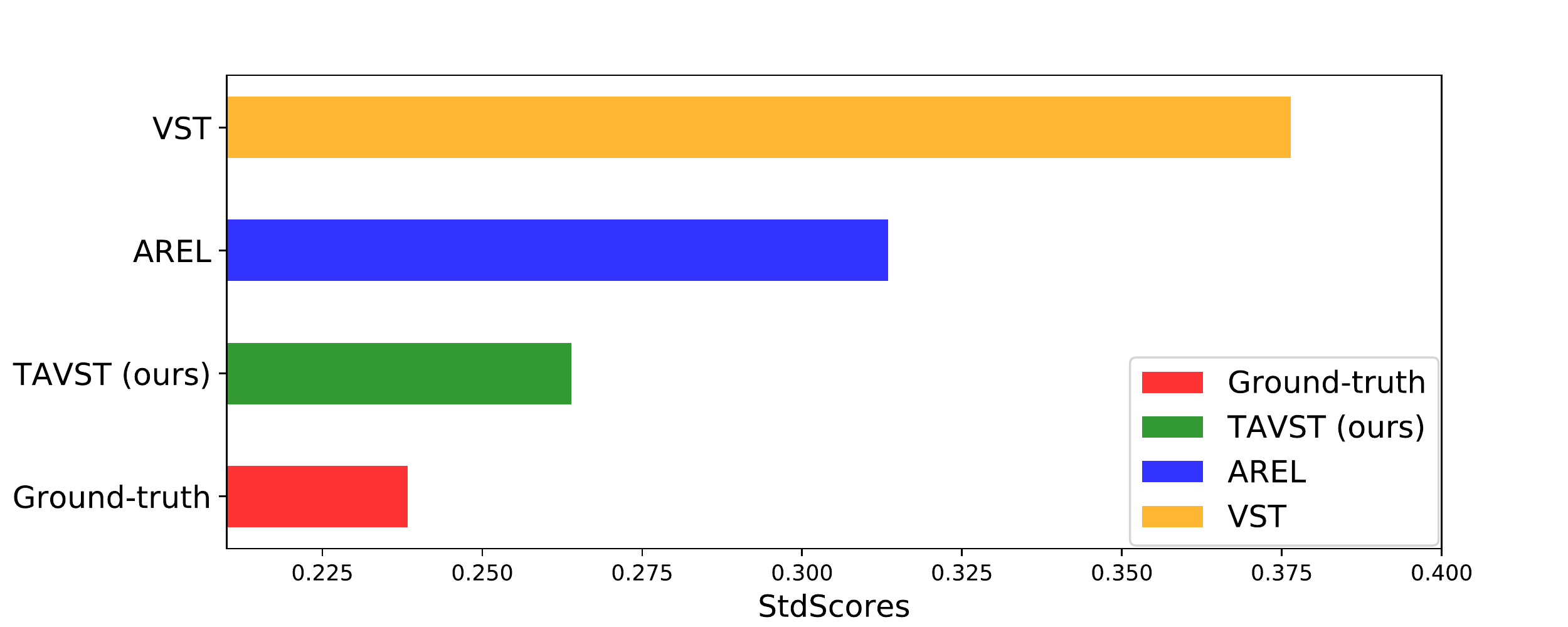}
  \vspace{-2mm}

  \caption{The comparison of in-story sentiment standard deviations among different methods.}
  \label{fig:in-story-sentiment}
\end{figure}

\paragraph{In-story Sentiment Consistency}
We argue that the sentiment of sentences in a story should be consistent given the album is related to a certain topic. For each story, we obtain a vector with 5 sentiment scores in correspondence to 5 sentences. We then calculate the standard deviation for the vector to represent the divergence score of a story. For each model, we average the divergence scores of all stories generated as its final score. Figure \ref{fig:in-story-sentiment} presents the results from different models. Results illustrate that our method can generate stories with higher in-story sentiment consistency.

\begin{wraptable}{r}{0.5\textwidth}

\centering
\small
\begin{tabular}{l|cc|cc}
\hline
\multirow{3}{*}{Method} & \multicolumn{2}{c|}{Positive}                     & \multicolumn{2}{c}{Negative}           \\ \cline{2-5} 
                        & new year's & sporting & breaking & car  \\
                        & eve & event & up & accident \\ 
\hline
GT            & 1.20                      & 1.25                  & 0.48             & 0.38              \\
TAVST                   & 1.31                      & 1.17                  & 0.53             & 0.42              \\
AREL                    & 1.22                      & 1.15                  & 0.95              & 0.81               \\
VST                     & 1.65                      & 1.82                  & 1.66              & 1.84               \\ \hline
\end{tabular}
\smallskip
\vspace{-1mm}

\caption{Sentiment scores corresponding to different types of events. Note that higher score indicates more positive polarity.}
\vspace{-2mm}

\label{tab:sentiment_consistence_with_topic}

\end{wraptable}

\paragraph{Topic-Story Sentiment Consistency}
Considering that albums related to some events might express a tendency to a certain polarity. For example, the sentiment of stories about \emph{new year's eve} are more likely to be positive while the sentiment of stories about \emph{breaking up} are more likely to be negative. We enumerate albums with different event types to see if the model has the ability to generate stories with sentiment consistent with the type of events. For each story, we add all the sentiment scores of sentences as its final score. The higher score a story obtain, the more positive the story is. Four types of events are considered. Results are shown in Table~\ref{tab:sentiment_consistence_with_topic}. In general, all automatic models tend to generate stories with higher sentiment scores compared to human-written stories. This is because a large portion of albums in the dataset are related to positive events.  Both ~\emph{VST} and~\emph{AREL} generate stories with similar sentiment scores for both types of events. This indicates that they are not able to distinguish positive and negative events. With the guidance of topic description, our model \emph{TAVST} is able to distinguish events with different sentiment tendency. 

\begin{figure*}[t!]
\begin{center}
\includegraphics[width=\textwidth]{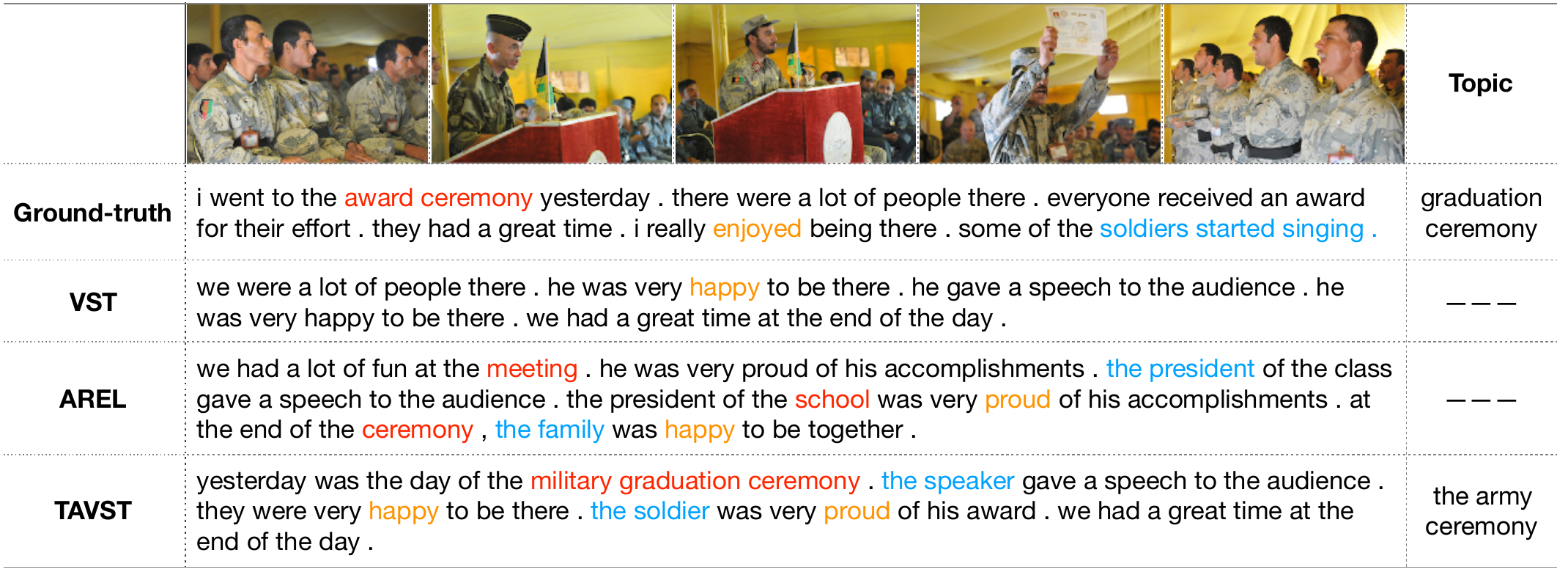}  
\end{center}
\vspace*{-3ex}
   \caption{Story samples generated by different models and the ground truth. The words in red, blue and yellow color represent the topic, subject, and emotion, respectively. (Best viewed in color)}
   \label{fig:case_1}
\vspace*{-2ex}
\end{figure*}

\subsection{Case Study}
Figure \ref{fig:case_1} shows an example of the ground-truth story and stories generated automatically by different models. The words in red, blue and yellow color represent the topic, subject, and emotion, respectively. Our model shows promising results according to topic consistency, which further confirms that our model can extract appropriate topic which serves as the guidance of generating a topic-consistent story.

\section{Related work}

This paper is related to the fields of image captioning, visual storytelling and multi-task learning. 

\paragraph{Image Captioning} 
In early works \citep{Yang2011CorpusGuidedSG,Elliott2013ImageDU}, image captioning is treated as a ranking problem, which is based on retrieval models to identify similar captions from the database. Later, the end-to-end frameworks based on the CNN and RNN are adopted by researchers~\citep{xu2015show,karpathy2015deep,vinyals2017show,dai2017towards}.
Such works focus on the literal description of image content, while the generated texts is limited in a single sentence. 

\paragraph{Visual Storytelling} Visual storytelling is the task of generating a narrative paragraph for an image stream. 
\citet{huang2016visual} introduce the first dataset (VIST) for visual storytelling, and establishes some baseline approaches. 
An attention-based RNN with a skip gated recurrent unit \citep{liu2016storytelling} is designed to maintain longer range information. 
In addition, \citet{yu2017hierarchically} design a hierarchically-attentive RNN structure. 
More recently, \citet{wang2018show} and \citet{Wang:2018tda} propose to utilize reinforcement learning frameworks for this task. \citet{wang2020storytelling} propose to translate images to graph-based semantic representations to benefit representing images, while it is not fair to compare with our method because they introduce information from other datasets.
The most similar work to ours is from~\citet{huang2019hierarchically}. They propose to generate a local semantic concept for each image in the sequence and generate a sentence for each image using a semantic compositional network in a fashion of hierarchical reinforcement learning. Although both of us consider topic information to facilitate the story generation. Our model is different from three aspects. First, the concepts of topic are different. We treat topic as the global semantic context of the album while topic represents local semantic information in their case adhering to each single image. Second, our modeling topic is more interpretable. We generate topic description directly instead of producing latent representation and this provides more insights for further improving the performance. Third, the communication framework is compatible with any RL based training methods. Experiment results also show that with RL, our framework can outperform their model. 

\paragraph{Multi-Task Learning}
\citet{collobert2008unified} first propose a method for processing NLP tasks in a deep learning framework using multi-task learning. \citet{Jing2018OnTA} build a multi-task learning framework which jointly performs the prediction of tags and the generation of paragraphs. These multi-task learning methods share a certain network structure, and at the output layer design a specific network structure for different tasks, improving the performance of different tasks. However, unlike these multi-task learning methods, we use another multi-agent method \citep{sukhbaatar2016learning,wang2019multi}. In this work, we define two kinds of agents for two generation tasks which can interact and share useful information. We also notice that in other areas, there are also some works \citep{xing2017topic,wang-etal-2019-topic-aware} consider incorporating topic information. 

\section{Conclusions and Future Work}
In this paper, we introduce a topic-aware visual storytelling task, which identifies the global semantic context of a given image sequence and then generate the story with the help of such topic information. We propose a multi-agent communication framework that combines two generation tasks namely topic description generation and story generation effectively. In future, we will explore to model topic generation as a keyword extraction task.

\section*{Acknowledgements}
This work is partically supported by National Natural Science Foundation of China (No. 71991471), Science and Technology Commission of Shanghai Municipality Grant (No.20dz1200600, No.18DZ1201000, 17JC1420200).

\bibliographystyle{coling}
\bibliography{coling2020}

\clearpage
\newpage
\appendix

\section*{Supplementary Material}
\label{sec:supp}



\section{Hyper-Parameter Sensitivity Analysis}
\label{parameter_analysis}
\paragraph{Impact of $\lambda_1$ and $\lambda_2$} In our experiments, we try many different values of $\lambda_1$ and $\lambda_2$ on the validation set. We found that $\lambda_1$ and $\lambda_2$ = 0.7 best balances the topic extractor and the story generator. As we know, the overall performance of our model depends more on the generated story rather than the generated topic. The topic guides story generation, which plays an auxiliary role. So the weight for the topic extractor should be lower than the weight of story generator.

\paragraph{Impact of $N_{iter}$ and $\beta$} The choice of the hyper-parameter $N_{iter}$ affects the performance of the model. In our experiments, we find that when $N_{iter}$ = 1 or 2, the performance of the model is better than $N_{iter}$ = 0, and $N_{iter}$ = 2 performs best on the validation set. But when $N_{iter}$ = 3, the performance declines. In addition, we observe that $\beta$ = 0.3 plays a very good regulating effect.

\paragraph{Impact of $\alpha$}
The hyper-parameter $\alpha$ controls the trade-off between MLE and RL objectives. For comparison, we set $\alpha$ to be [0, 0.5, 0.8, 1] in our experiments. The results show that, when $\alpha$ =[0.5,0.8,1], the model achieves a better performance than $\alpha$ = 0; and $\alpha$ =0.8 best balances the RL loss and the MLE loss. 

\section{The structure of Co-Attention Encoding Module}

\begin{figure}[h]
\begin{center}
\includegraphics[width=0.5\textwidth]{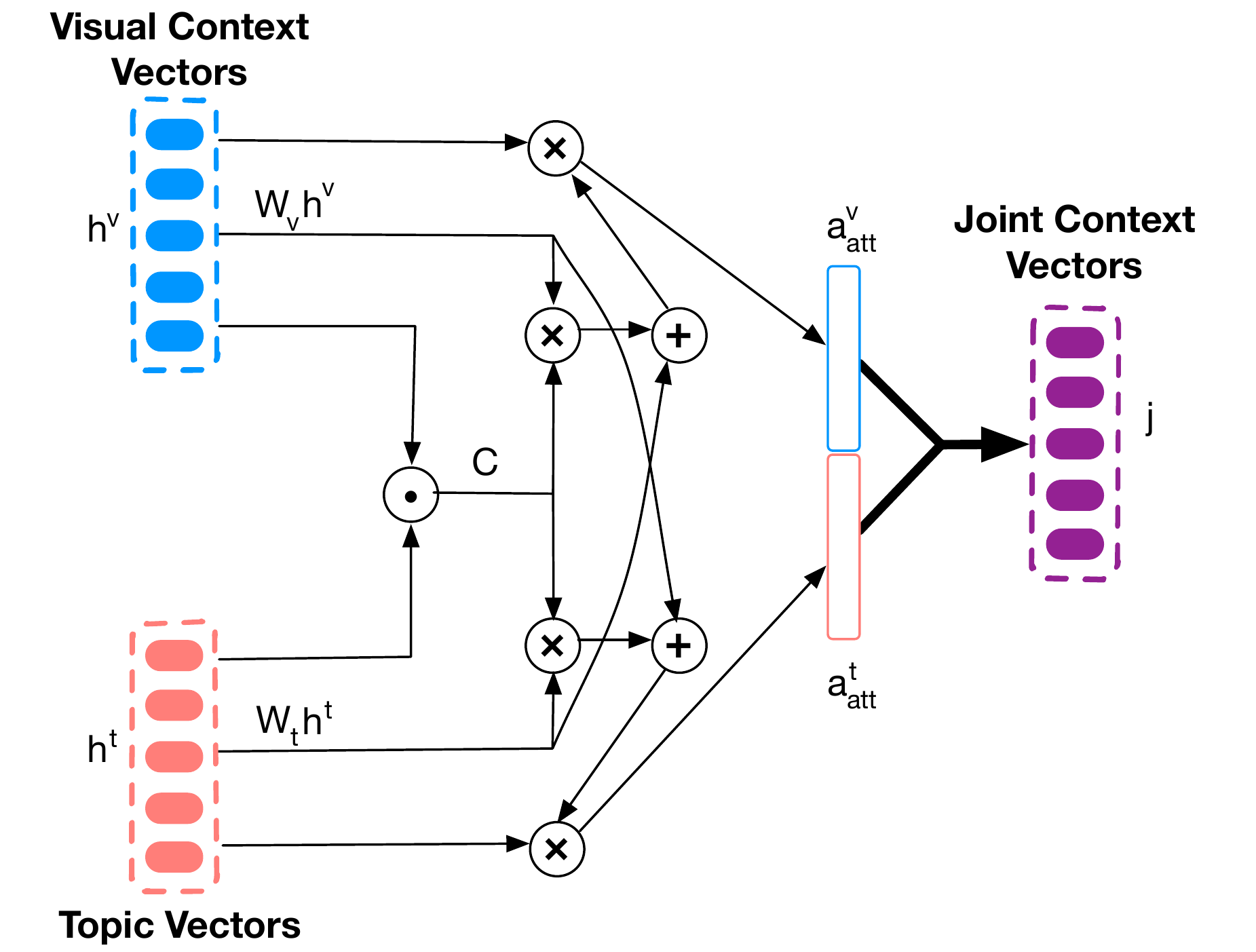}  
\end{center}
  \caption{Co-attention mechanism to combine visual context vector and topic vector.}
  \label{fig:co-attention}
\end{figure}

\section{Surveys for Human Evaluation}

\begin{figure}
	\centering
	\includegraphics[width=1.0\textwidth]{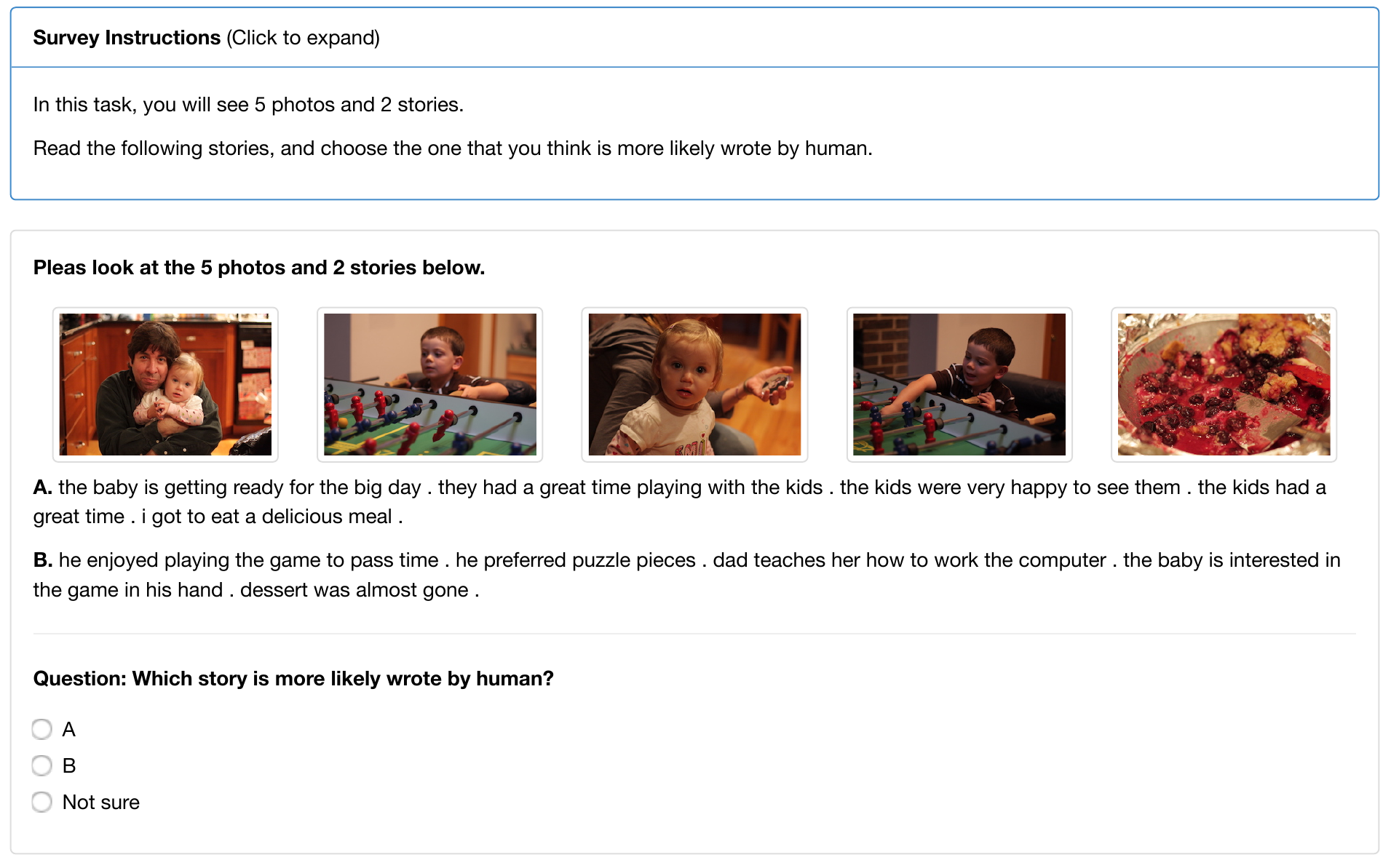} 
	\caption{Turing Test Survey}
	\label{fig:turing-test}
\end{figure}

\begin{figure}
	\centering
	\includegraphics[width=1.0\textwidth]{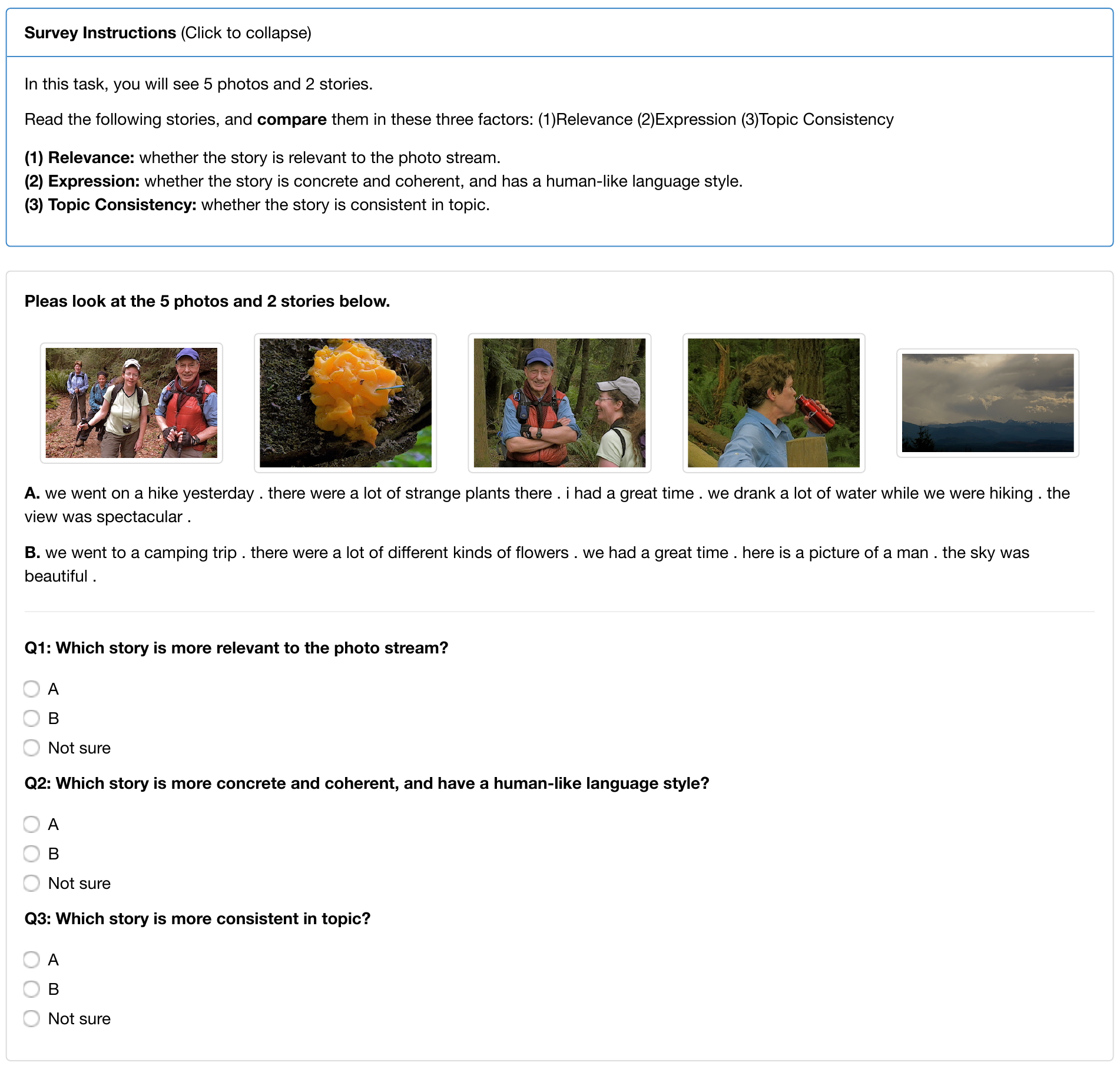} 
	\caption{Pairwise Comparison Survey}
	\label{fig:pairwise-comparison}
\end{figure}

\end{document}